%% file: 1819.tex
\documentclass[runningheads]{llncs}
\usepackage{graphicx}
\usepackage{subfigure}
\usepackage{comment}
\usepackage{amsmath,amssymb} 
\usepackage{color}

\usepackage{tabularx}
\usepackage{hyperref}
\usepackage{url}

\hypersetup{hidelinks} 

\begin{document}
\pagestyle{headings}
\mainmatter
\def\ECCVSubNumber{1819}  
\title{Event Enhanced High-Quality Image Recovery}

\titlerunning{Event Enhanced High-Quality Image Recovery}
%
\author{Bishan Wang \inst{*} \and
	Jingwei He\inst{*} \and
	Lei Yu\inst{\dagger} \and
	Gui-Song Xia \and
	Wen Yang }
%
\authorrunning{B. Wang et al.}
%
\institute{Wuhan University, Wuhan, China\\
	\email{\{wangbs,jingwei\_he,ly.wd,guisong.xia,yangwen\}@whu.edu.cn}}
\maketitle

\begin{abstract}
With extremely high temporal resolution, event cameras have a large potential for robotics and computer vision. However, their asynchronous imaging mechanism often aggravates the measurement sensitivity to noises and brings a physical burden to increase the image spatial resolution. To recover high-quality intensity images, one should address both denoising and super-resolution problems for event cameras. Since events depict brightness changes, with the enhanced degeneration model by the events, the clear and sharp high-resolution latent images can be recovered from the noisy, blurry and low-resolution intensity observations. Exploiting the framework of sparse learning, the events and the low-resolution intensity observations can be jointly considered. Based on this, we propose an explainable network, an event-enhanced sparse learning network ({\bf eSL-Net}), to recover the high-quality images from event cameras. After training with a synthetic dataset, the proposed eSL-Net can largely improve the performance of the state-of-the-art by {\bf 7-12 dB}. Furthermore, without additional training process, the proposed eSL-Net can be easily extended to generate continuous frames with frame-rate as high as the events. 	
 
\end{abstract}

\begin{figure}[t]	
	\centering	
	\includegraphics[width=12cm]{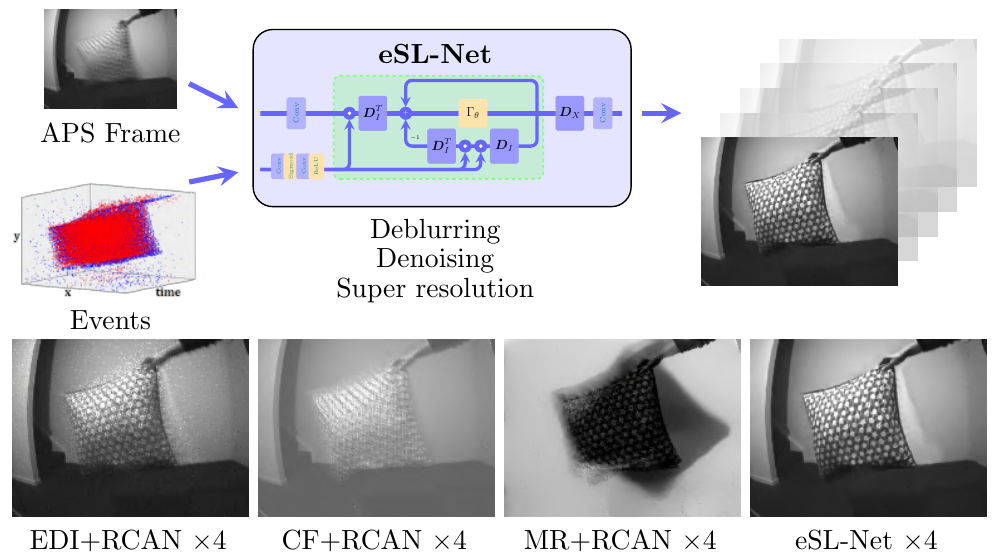}	
	\caption{Our eSL-Net reconstructs high-resolution, sharp and clear intensity images for event cameras by APS frames and the corresponding event sequences. The eSL-Net performs much better than EDI \cite{pan2019bringing}, CF \cite{scheerlinck2019a} and MR \cite{munda2018} superimposing a SR network RCAN \cite{zhang2018image}.}	
	\label{highlight}
\end{figure}

\section{Introduction}

\let\thefootnote\relax\footnotetext{$\dagger$ Corresponding Author}
\let\thefootnote\relax\footnotetext{* Equal contribution}

Unlike the standard frame-based cameras, the event camera is a bio-inspired sensor that produce asynchronous ``events'' with very low latency (1 $\mu$s), leading to extremely high temporal resolution \cite{lichtsteiner2008128,liu2010a,posch2011,brandli2014b,son2017}. Naturally, it is immune to motion blurs and has  highly appealing promise for low/high-level vision tasks \cite{pan2019bringing,almatrafi2019a,vidal2018a}. However, the generated event streams can only depict the scene changes instead of the absolute intensity measurements. Meanwhile, the asynchronous data-driven mechanism also prohibits directly applying existing algorithms designed for standard cameras to event cameras. Thus the high-quality intensity image reconstruction from event streams is essentially required for visualization and provides great potentials to bridge the event camera to many high-level vision tasks that have been solved with standard cameras \cite{kim2014,kim2016,munda2018,rebecq2019,gehrig2019a}.

In order to achieve the low latency property, event cameras capture brightness changes of each pixels independently \cite{brandli2014b,gallego2019}. This mechanism aggravates the measurement sensitivity to noises and brings a physical burden to increase the image spatial resolution. Thus, the recovering of high-quality images from event cameras is a very challenge problem, where the following issues should be addressed simultaneously.

\begin{itemize}
	\item {\bf Low frame-rate and blurry intensity images:} The APS (Active Pixel Sensor) frames are with relatively low frame-rate ($\geq 5\ ms$ latency). And the motion blur is inevitable when recording highly dynamic scenes.
	\item {\bf High level and mixed noises:} The thermal effects or unstable light environment can produce a huge amount of noisy events. Together with the noises from APS frames, the reconstruction of intensity image would fall into a mixed noises problem.
	\item {\bf Low spatial-resolution:} The leading commercial event cameras are typically with very low spatial-resolution. And there is a balance between the spatial-resolution and the latency.
\end{itemize}

To address the problem of noises for recovering images from event cameras, various methods have been proposed. Barua et. al \cite{barua2016} firstly proposed a learning-based approach to smooth the image gradient by imposing sparsity regularization, then exploited Poisson integration to recover the intensity image from denoised image gradient.  Instead of sparsity, Munda et. al \cite{munda2018} introduced the manifold regularization imposed on the event time surface and proposed a real-time intensity reconstruction algorithm. With these hand-crafted regularizations, the noises can be largely alleviated, however, some artifacts (e.g. blurry edges) are meanwhile produced. 
Recent works turn to convolutional neural network (CNN) for event-based intensity reconstruction, where the network is trained end-to-end with paired events and intensity images \cite{rebecq2019,wang2019,gehrig2019a,i2019}. Implicitly, the convolutional kernels with trained parameters are commonly able to reduce the noises. However, the man-made networks are often lack of physical mean and thus difficult to deal with both events and APS frames \cite{i2019}.

Besides the noise issue, a super-resolution algorithm is also urgent at present phase to further improve intensity reconstructions for high-level vision tasks, e.g. face recognition, but few of progress has been made in this line yet. Even though one can apply existing super-resolution algorithms to the low-resolution intensity frames (reconstructed), a comprehensive approach will be more desirable.

To the best of our knowledge, few study is able to simultaneously resolve all above three tasks, leaving an open problem: \textit{Is it possible to find a unified framework to consider denoising, debluring and super-resolution simultaneously?} To answer this question, we propose to employ a powerful tool \textit{sparse learning} to address the three tasks. General degeneration model for blurry images with noises and low-resolution, often assumes the whole image shares the same blurring kernel. However, events record intensity changes at a very high temporal resolution, which can enhance the degeneration model effectively to represent motion blur effect. The enhanced degeneration model provides a road to recover HR sharp and clear latent images from APS frames and their event sequences. We can solve the model by casting it to the framework of sparse learning, which also leads to its natural ability to resist noise. In this paper, we propose the {\bf eSL-Net} to recover high-quality images for event cameras. Specially, the eSL-Net trained by our synthetic dataset can be generalized to real scenes and without additional training process the eSL-Net can be easily extended to generate high frame-rate videos by transforming the event sequence. Experimental results show the proposed eSL-Net can improve high-quality intensity reconstruction.

Overall, our contributions are summarized as below:
\begin{itemize}
	\item  We propose an event enhanced degeneration model for the high-quality image recovery based on event cameras. Based on this, exploiting the framework of sparse learning, we propose an explainable network, an event-enhanced sparse learning network (eSL-Net), to recover the high-quality images from event cameras.  
	\item Without retraining process, we propose an easy method to extend the eSL-Net for high frame-rate and high-quality video recovery.
	\item We build a synthetic dataset for event camera to connect events, LR blurry images and the HR sharp clear images.
\end{itemize}

Dataset, code, and more results are available at: \url{https://github.com/ShinyWang33/eSL-Net}.

\section{Related Works}

{\bf Event-based Intensity Reconstruction:}
Early attempts of reconstructing intensity from pure events are commonly based on the assumption of brightness constancy, i.e. static scenes \cite{kim2014}. The intensity reconstruction is then addressed by simultaneously estimating the camera movement, optical flow and intensity gradient \cite{kim2016}. In \cite{cook2011}, Cook et al. propose a bio-inspired and interconnected network to simultaneously reconstruct intensity frames, optical flow and angular velocity for small rotation movements. Later on, Bardow et. al \cite{bardow2016} formulate the intensity change and optical flow in a unified variational energy minimization framework. By optimization, one can simultaneously reconstruct the video frames together with the optical flow. On the other hand, another research line on intensity reconstruction is the direct event integration method \cite{scheerlinck2019a,munda2018,pan2019bringing}, which does not rely on any assumption about the scene structure or motion dynamics.

While the APS frames contain relatively abundant textures, events and APS frames can be used as complementary sources for event-based intensity reconstruction. In \cite{scheerlinck2019a}, events are approximated as the time differential of intensity frames. Based on this, a complementary filter is proposed as a fusion engine and nearly continuous-time intensity frames can be generated. Pan et. al \cite{pan2019bringing} have proposed an event-based deblurring approach by relating blurry APS frames and events with an event-based double integration (EDI) model. Afterwards, a multiple-frame EDI model is proposed for high-rate video reconstruction by further considering frame-to-frame relations \cite{pan2019a}.


{\bf Event-based Super-resolution:} Even though event cameras have extremely high temporal frequency, the spatial (pixel) resolution is relative low and yet not easy to be resolved physically \cite{gallego2019}. Few of progress has been made to event-based super-resolution. To the best of our knowledge, only one very recent work \cite{i2019}, called SRNet, has been released when we are preparing this manuscript. Comparing to SRNet, our proposed approach differs in the following aspects: (1) we proposed a unified framework to simultaneously resolve the tasks including denoising, deblurring and superresolution, while SRNet \cite{i2019} cannot directly deal with blurring or noisy inputs; (2) the proposed network is completely interpretable with meaningful intermediate processes; (3) our framework reconstructs the intensity frame by fusing events and APS frames, while SRNet is proposed for reconstruction from pure events.

\section{Problem Statement}
\subsection{Events and Intensity Images}

Event camera triggers events whenever the logarithm of the intensity changes over a pre-setting threshold $c$, 

\begin{equation}\label{con:event_trigger}
 \log(\boldsymbol{I}_{xy}(t))-\log(\boldsymbol{I}_{xy}(t-\Delta t)) = p \cdot c
\end{equation}
where $\boldsymbol{I}_{xy}(t)$ and $\boldsymbol{I}_{xy}(t-\Delta t)$ denote the instantaneous
intensities at time $t$ and $t-\Delta t$ for a specific pixel location $(x, y)$, $\Delta t$ is the time since the last event at this pixel location, $p \in \left\lbrace +1,-1 \right\rbrace $ is the polarity representing the direction (increase or decrease) of the intensity change. Consequently, an event is made up of $(x,y, t, p)$.

In order to facilitate expression of events, for every location $(x,y)$ in the image, we define $e_{xy}(t)$ as a function of continuous time $t$ such that:
\begin{equation}\label{event_with_dirac}
e_{xy}(t) \triangleq  p \delta(t-t_0)
\end{equation}
whenever there is an event $ (x,y,t_0,p) $. Here, $\delta(\cdot)$ is the Dirac function \cite{dirac1981principles}. As a result, a sequence of discrete events is turned into a continuous time signal.

In addition to event sequence, many event cameras \textit{e.g.}, DAVIS \cite{brandli2014b}, can provide grey-scale intensity images simultaneously with slower frame-rate. And mathematically, the $f$-th frame of the observed intensity image $\boldsymbol{Y}[f]$ during the exposure interval $[t_f,t_f+T]$ could be modeled as an average of sharp clear latent intensity images $\boldsymbol{I}(t)$ \cite{pan2019bringing}:
\begin{equation}\label{blur_model}
\boldsymbol{Y}[f]=\frac{1}{T} \int_{t_f}^{t_f+T} \boldsymbol{I}(t) d t
\end{equation}
Suppose that $\boldsymbol{I}_{xy}(t_r)$ is the sharp clear latent intensity image at any time $t_r\in [t_f,t_f+T]$, we have the following relationship according to   \eqref{con:event_trigger} and   \eqref{event_with_dirac},
$\log\left(\boldsymbol{I}_{xy}(t)\right)=\log\left(\boldsymbol{I}_{xy}(t_r)\right)+c \int_{t_r}^{t} e_{xy}(s) d s$, then
\begin{equation}\label{event_image_clear}
\boldsymbol{Y}_{xy}[f]=\frac{\boldsymbol{I}_{xy}(t_r)}{T} \int_{t_f}^{t_f+T} \exp \left(c \int_{t_r}^{t} e_{xy}(s) d s\right) d t
\end{equation}
Since each pixel can be treated separately, subscripts $x$, $y$ are often omitted henceforth. Finally, considering the whole pixels, we can get a simple model connecting events, the observed intensity image and the latent intensity image:
\begin{equation}\label{mathmodel}
\boldsymbol{Y}[f]= \boldsymbol{E}(t_r) \odot \boldsymbol{I}(t_r)
\end{equation}
with $\boldsymbol{E}(t_r) =\frac{1}{T} \int_{t_f}^{t_f+T} \exp (c \int_{t_r}^{t} e(s) d s) d t$ being double integral of events at time $t_r$ \cite{pan2019bringing} and $\odot$ denoting the Hadamard product. 

\subsection{Event Enhanced Degeneration Model}

Practically, the non-ideality of sensors and the relative motion between cameras and target scenes may largely degrade the quality of the observed intensity image $\boldsymbol{Y}[f]$ and make it noisy and blurry. Moreover, even though event cameras have extremely high temporal resolution, the spatial pixel resolution is relatively low due to the physical limitations. With these considerations, \eqref{mathmodel} becomes:
\begin{equation}\label{general_model}
\begin{split}
\boldsymbol{Y}[f] &= \boldsymbol{E}(t_r) \odot \boldsymbol{ I}(t_r) +\boldsymbol{\varepsilon} \\
\boldsymbol{ {I}}(t_r) &= \boldsymbol{P}\boldsymbol{X}(t_r)
\end{split}
\end{equation}
with $ \boldsymbol{\varepsilon} $ the measuring noise which can be assumed to be white Gaussian, $\boldsymbol{P}$ the downsampling operator and $\boldsymbol{X}(t_r)$ the latent clear image with high-resolution (HR) at time $t_r$. Consequently,   \eqref{general_model} is the degeneration model where events are exploited to introduce the motion information. 

Given the observed image $\boldsymbol{Y}[f]$, the corresponding triggered events and the specified time $t_r\in [t_f,t_f+T]$, our goal is to reconstruct a high quality intensity image $\boldsymbol{X}$ at time $t_r$. Obviously, it is a multi-task and ill-posed problem where denoising, deblurring and super-resolution should be addressed simultaneously. 

In the following, we will first address the problem of reconstructing single high quality intensity image from events and a degraded LR blurry image. Then, the method to extend to generate high frame-rate video is addressed in Section 5.

\section{Event Enhanced High-Quality Image Recovery}
\subsection{Event-Enhanced Sparse Learning}
Many methods were proposed for image denoising, deblurring and SR~\cite{dncnn2017,elad2006image,pan2016blind,srcnn2015}. However, most of them can not be applied for event cameras directly due to the asynchronous data-driven mechanism. Thanks to the sparse learning, we could integrate the events into sparsity framework and reconstruct satisfactory images to solve the aforementioned problems.

In this section, the expression of the time $t_r$ and frame index $f$ is temporally removed for simplicity. Then we arrange the image matrices as column vectors, \textit{i.e.}, $ \boldsymbol{Y} \in \mathbb{R}^{N \times 1} $, $ \boldsymbol{ I} \in \mathbb{R}^{N \times 1} $, $ \boldsymbol{\varepsilon} \in \mathbb{R}^{N \times 1} $ and $ \boldsymbol{X} \in \mathbb{R}^{sN \times 1} $, thus the blurring operator can be represented as $\boldsymbol{E}=\text{diag}(e_1,e_2,\dots,e_N) \in \mathbb{R}^{N \times N}$, where $e_1,e_2,\dots,e_N$ are the elements of original blurring operator, so does $\boldsymbol{P} \in \mathbb{R}^{N \times sN}$, where $ s $ denotes the downsampling scale factor and $N$ denotes the product of height $H$ and width $W$ of the observed image $\boldsymbol{Y}$. Then, according to   \eqref{general_model}, we have:
\begin{equation}
\begin{split}
\boldsymbol{Y} &= \boldsymbol{ E} \boldsymbol{ I} +\boldsymbol{\varepsilon} \\
\boldsymbol{ {I}} &= \boldsymbol{P}\boldsymbol{X}
\end{split}
\end{equation}

The reconstruction from the observed image $ \boldsymbol{Y} $ to HR sharp clear image $\boldsymbol{X}$ is highly ill-posed since the inevitable loss of information in the image degeneration process. Inspired by the success of Compressed Sensing~\cite{cs2006}, we assume that LR sharp clear image $ \boldsymbol{ I} $ and HR sharp clear image $ \boldsymbol{X} $ can be sparsely represented on LR dictionary $ \boldsymbol{D}_{{I}} $ and HR dictionary $ \boldsymbol{D}_{X} $, \textit{i.e.}, $ \boldsymbol{ I} = \boldsymbol{D}_{{I}} \boldsymbol{\alpha}_{ I} $ and $ \boldsymbol{X} = \boldsymbol{D}_{X} \boldsymbol{\alpha}_{ X} $ where $ \boldsymbol{\alpha}_{ I} $ and $ \boldsymbol{\alpha}_{ X} $ are known as sparse codes. Since the downsampling
operator $\boldsymbol{P}$ is linear, LR sharp clear image $ \boldsymbol{I} $ and HR sharp clear image $ \boldsymbol{X} $ can share the same sparse code, \textit{i.e.} $ \boldsymbol{\alpha} = \boldsymbol{\alpha}_{ I} = \boldsymbol{\alpha}_{ X}$ if the dictionaries $ \boldsymbol{D}_{{I}} $ and $ \boldsymbol{D}_{X} $ are defined properly. Therefore, given an observed image $ \boldsymbol{Y} $, we first need to find its sparse code on $ \boldsymbol{D}_{{I}} $ by solving the LASSO \cite{tibshirani1996regression} problem below:
\begin{equation} \label{ev-lasso}
\arg \min _{{\boldsymbol{\alpha}}} \frac{1}{2}\|\boldsymbol{Y}-{\boldsymbol{{E}}\boldsymbol{D}_{{I}}}{\boldsymbol{\alpha}}\|_{2}^{2}+\lambda\|{\boldsymbol{\alpha}}\|_{1}
\end{equation}
where $ \|\cdot\|_p $ denotes the $\ell_p$-norm, and $\lambda$ is a regularization coefficient.

To solve   \eqref{ev-lasso}, a common approach is to use iterative shrinkage thresholding algorithm (ISTA) \cite{daubechies2004an}. At the $ n $-th iteration, the sparse code is updated as:
\begin{equation}\label{ev-ista}
\begin{split}
\boldsymbol{\alpha}_{n+1}&=\Gamma_{\frac{\lambda}{L}}(\boldsymbol{\alpha}_n+{\frac{1}{L}}{( \boldsymbol{E}\boldsymbol{D}_I )}^T(\boldsymbol{Y}-{\boldsymbol{E}\boldsymbol{D}_I}{\boldsymbol{\alpha}_n})) \\
&=\Gamma_{\frac{\lambda}{L}}(\boldsymbol{\alpha}_n-{\frac{1}{L}}\boldsymbol{D}_I^T\boldsymbol{E}^T\boldsymbol{E}\boldsymbol{D}_I\boldsymbol{\alpha}_n+{\frac{1}{L}}\boldsymbol{D}_I^T\boldsymbol{E}^T\boldsymbol{Y})
\end{split}
\end{equation}
where $L$ is the Lipschitz constant, $ \Gamma_{\theta}(\beta) = \text{sign}(\beta)\text{max}(|\beta|-\theta,0) $ denotes the element-wise soft thresholding function. After obtaining the optimum solution of sparse code $ \boldsymbol{\alpha}^{*} $, we could finally recover HR sharp clear image $ \boldsymbol{X} $ by:
\begin{equation}\label{recovery-x}
\boldsymbol{X} = \boldsymbol{D}_{X}  \boldsymbol{\alpha}^{*}
\end{equation}
where $ \boldsymbol{D}_{X} $ is the HR dictionary.
\subsection{Network}
Inspired by \cite{gregor2010learning}, we can solve the sparse coding problem efficiently by integrating it into the CNN architecture. Therefore we propose an Event-enhanced Sparse Learning Net (\textbf{eSL-Net}) to solve problems of noise, motion blur and low spatial resolution in a unified framework.

The basic idea of eSL-Net is to map the update steps of event-based intensity reconstruction method to a deep network architecture that consists of a fixed number of phases, each of which corresponds to one iteration of   \eqref{ev-ista}. Therefore eSL-Net is an interpretable deep network.
\begin{figure}[t]	
	\centering	
	\includegraphics[width=12cm]{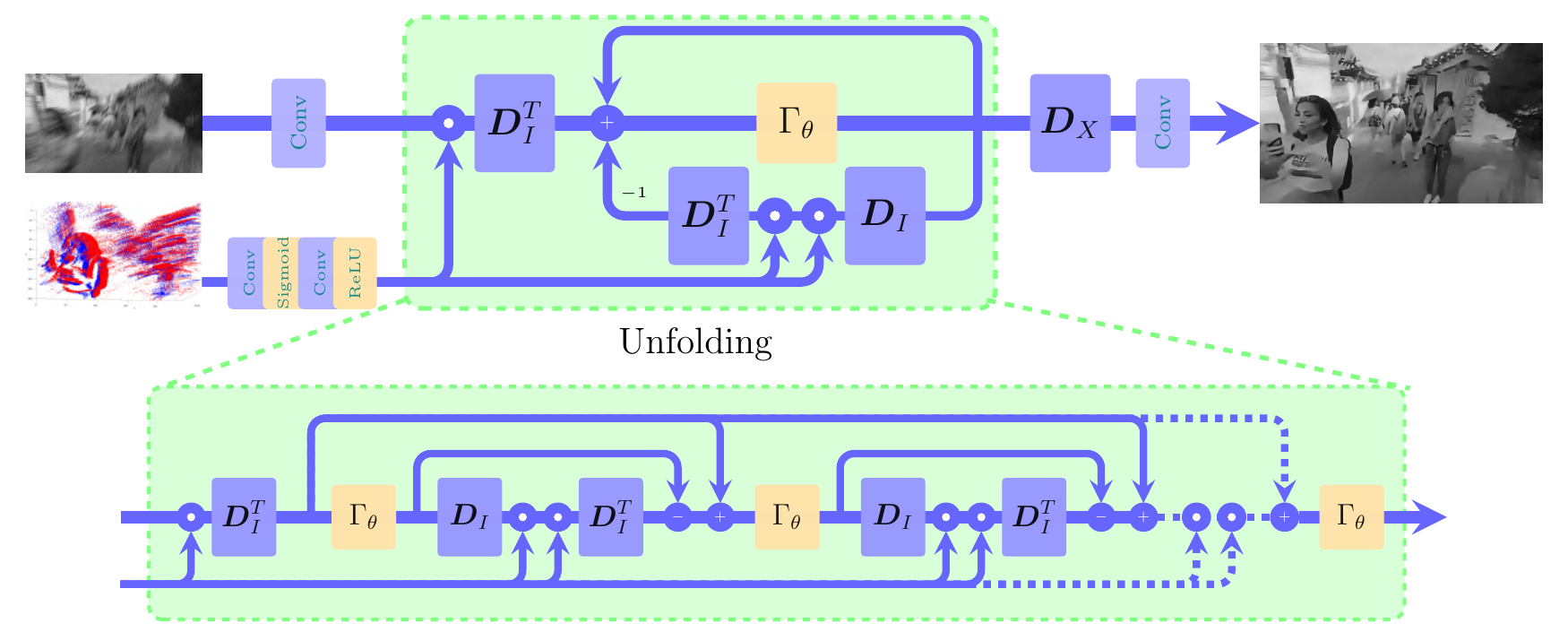}	
	\caption{eSL-Net Framework}	
	\label{framework}
\end{figure}

The whole eSL-Net architecture is shown as Fig. \ref{framework}. Obviously the most attractive part in the network is iteration module corresponding to   \eqref{ev-ista} in the green box. According to \cite{papyan2017convolutional}, when the coefficient in   \eqref{ev-ista} is limited to nonnegative, ISTA is not affected. It is easy to find the equality of the soft nonnegative thresholding operator $\Gamma_{\theta}$ and the ReLU activation function. We use ReLU layer to implement $\Gamma_{\theta}$. Convolution is a special kind of matrix multiplication, therefore we use convolution layers to implement matrix multiplication. Then the plus node in the green box with three inputs represents $\boldsymbol{\alpha}_n-{\frac{1}{L}}\boldsymbol{D}_I^T\boldsymbol{E}^T\boldsymbol{E}\boldsymbol{D}_I\boldsymbol{\alpha}_n+{\frac{1}{L}}\boldsymbol{D}_I^T\boldsymbol{E}^T\boldsymbol{Y}$ in \eqref{ev-ista}.

According to \eqref{mathmodel}, $\boldsymbol{E}$ is double integral of events. In discrete case, the continuous integral turns into discrete summation. More generally, we use the weighted summation, convolution, to replace integral. As a result, through two convolution layers with suitable parameters, the event sequence input 
can be transformed to approximative $\boldsymbol{E}$. What's more, convolution has some de-noise effect on event sequences.

Finally, the output of the iterative module, optimum sparse encoding $\boldsymbol{\alpha}^{*}$, is passed through a HR dictionary according to   \eqref{recovery-x}. In eSL-Net, we use convolution layers followed by shuffle layer to implement HR dictionary $\boldsymbol{D}_{X}$, due to the fact that the shuffle operator, arranging the pixels of different channels, can be regarded as a linear operator.

\begin{figure*}[t]
	\centering
	\subfigure[]{
		\begin{minipage}[t]{0.49\linewidth}
			\centering
			\includegraphics[width=0.98\linewidth]{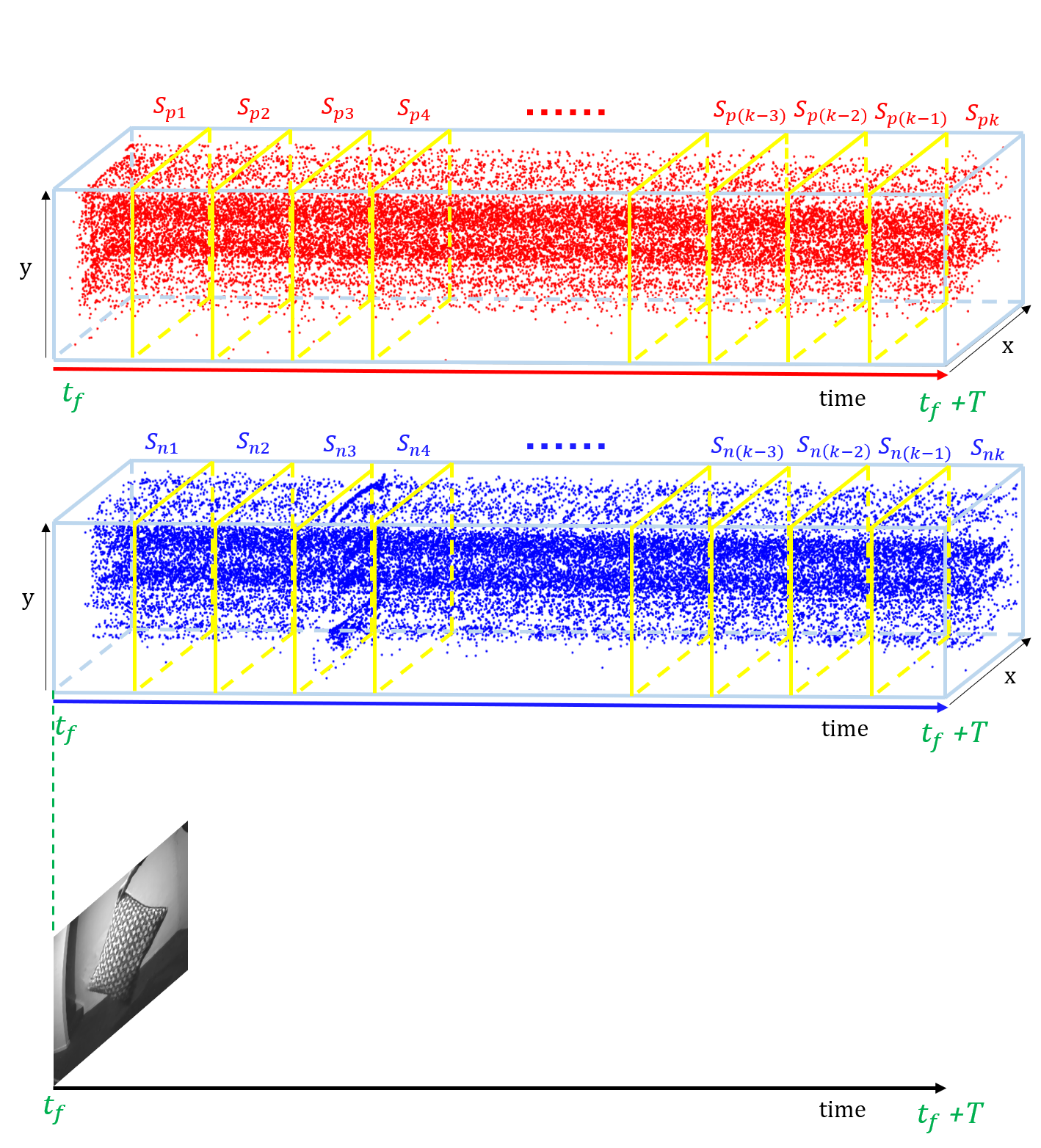}
		\end{minipage}%
	}\hspace*{-1.5mm}
	\subfigure[]{
		\begin{minipage}[t]{0.49\linewidth}
			\centering
			\includegraphics[width=0.98\linewidth]{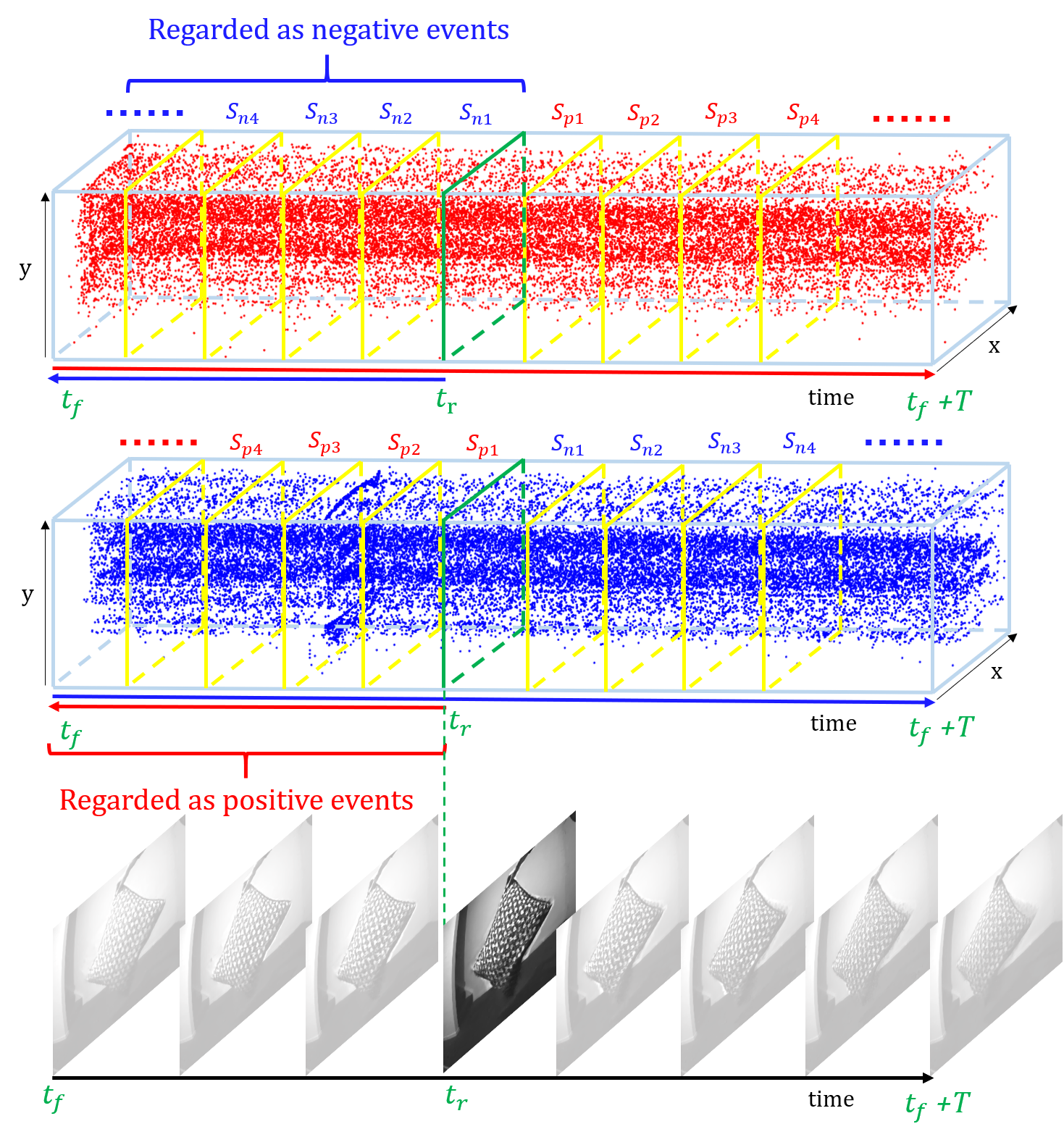}
		\end{minipage}%
	}
	\caption{(a) The method of transforming the event sequence to input event frames when eSL-Net is trained. The upper part is positive event sequence and the lower part is negative event sequence. (b) The method of transforming the event sequence to input event frames when eSL-Net outputs the latent frame at time $t_{r}$.}	
	\label{input event frame}
\end{figure*}

\subsection{Network Training}
 Because the network expects image-like inputs, we divide the time duration of the event sequence into $k$ equal-scale portions, and then $2k$ grayscale frames, $\boldsymbol{S_{pi}}(x,y)$, $\boldsymbol{S_{ni}}(x,y)$, $i=1,2, . ., k$ are formed respectively by merging the positive and negative events in each time interval, which is shown in Fig. \ref{input event frame} (a). $\boldsymbol{S_{pi}}(x,y)$ is the amount of positive events at $(x,y)$ and $\boldsymbol{S_{ni}}(x,y)$ is the amount of negative events at $(x,y)$. The input tensor (obtained by concatenating $\boldsymbol{Y}$, $\boldsymbol{S_{p1}}$, $\boldsymbol{S_{n1}}$, $\boldsymbol{S_{p2}}$, $\boldsymbol{S_{n2}}$, ..., $\boldsymbol{S_{pk}}$, $\boldsymbol{S_{nk}}$), of size $(1+2 \times k) \times H \times W$ is passed through eSL-Net and size of output is $1 \times sH \times sW$ ($s$ is upscale factor of the image).
 
 As shown in Fig. \ref{framework}, the eSL-Net is then fed with a pair of inputs including the $f$-th frame of the observed LR blurry and noisy intensity image $\boldsymbol{Y}[f]$ and its corresponding event sequence triggered between the time interval $[t_f,t_f+T]$. With such inputs, the output is the HR sharp and clear intensity image $\boldsymbol{X}$ at time $t_f$, as shown in Fig.~\ref{input event frame} (a). 

{\bf Loss :} we use $\ell_1$ loss which is a common loss in many image reconstruction methods. By minimizing $\ell_1$ loss, our network effectively learns to make the output closer to the desired image. And $\ell_1$ loss makes training process more stable.
\section{High Frame-Rate Video Generation}\label{high-frame}

With the $f$-th observed LR image frame $\boldsymbol{Y}[f]$ and the corresponding event sequence triggered during $[t_f,t_f+T]$, it is possible to reconstruct the latent intensity image $\boldsymbol{X}(t_f)$ by the trained eSL-Net, as shown in Fig.~\ref{input event frame} (a). Thus, the eSL-Net is trained for reconstructing the latent image at $t_f$.
\[
\boldsymbol{X}(t_f) = \mbox{eSL-Net}\left(\boldsymbol{Y}[f],\boldsymbol{E}(t_f)\right)
\]

In order to reconstruct the latent intensity image $\boldsymbol{X}(t_r)$ of $t_r\neq t_f$, one should get the learned double integral of events $\boldsymbol{E}(t_r)$ at time $t_r$. Let us consider the \eqref{event_image_clear}, the definition of the double integral implies that all events $e_{xy}(s)$ are fed into the network keeping the polarity and the order unchanged when $t_r=t_f$, but when $t_r \neq t_f$ the polarity and the order of the input events with timestamp less than $t_r$ should be reversed. It is worth noting that this reversion has special physical mean for event cameras where the polarity and order of the produced events respectively represent the direction and the relative time of the brightness change. So when reconstructing the latent intensity image $\boldsymbol{X}(t_r)$ with the trained network for $\boldsymbol{X}(t_f)$, we need to re-organize the input event sequence to match the pattern of the polarity and the order as in the training phase.

Consequently, instead of retraining the network, we propose a very simple preprocessing step for the input event sequence to reconstruct the latent intensity image $\boldsymbol{X}(t_r)$ for any $t_r\in [t_f,t_f+T]$. As shown in Fig.~\ref{input event frame} (b), the preprocessing step is only reversing the polarities and the orders of events with timestamp less than $t_r$. After that, the resulted new event sequence is then merged into frames as the inputs of eSL-Net. Theoretically, we can generate a video with frame-rate as high as the DVS's (Dynamic Vision Sensor) eps (events per second).

\section{Dataset Preparation}\label{dataset}
In order to train the proposed eSL-Net, a mass of LR blurry noisy images with corresponding HR ground-truth images and event sequences are required. However, there exists no such large-scale dataset. This encourages us to synthesize a new dataset with LR blurry noisy images and the corresponding HR sharp clear images and events. And Section \ref{exper} shows that, although trained on synthetic data, the eSL-Net is able to be generalized to real-world scenes \cite{pan2019bringing}. 

{\bf HR clear images:} We choose the continuous sharp clear images with resolution of $1280 \times 720$ from GoPro dataset \cite{nah2019ntire} as our ground truth.

{\bf LR clear images:} LR sharp clear images with resolution of $320 \times 180$ are obtained by sampling HR clear images with bicubic interpolation, that are used as ground truth for the eSL-Net without SR.

{\bf LR blurry images:} The GoPro dataset \cite{nah2019ntire} also provides LR blurry images, but we have to regenerate them due to the ignorance of exposure time. Mathematically, during the exposure, a motion blurry image can be simulated by averaging a series of sharp images at a high frame rate \cite{Nah2017Deep}. However, when the frame rate is insufficient, e.g. $120$ fps in GoPro \cite{nah2019ntire}, simple time averaging would lead to unnatural spikes or steps in the blur trajectory~\cite{Wieschollek2017Learning}. To avoid this issue, we first increase the frame-rate of LR sharp clear images to $ 960 $ fps by the method in \cite{Niklaus2017VideoFI}, and then generate LR blurry images by averaging $ 17 $ continuous LR sharp clear images. Besides, to better simulate the real situation, we add additional white Gaussian noise with standard deviation $\sigma=4$ ($\sigma=4$ is the approximate mean of the standard deviations of many smooth patches in APS frames in the real dataset) to the LR blurry images.

{\bf Event sequence:} To simulate events, we resort to the open ESIM \cite{rebecq2018esim} which can generate events from a sequence of input images. For a given LR blurry image, we input the corresponding LR sharp clear images ($ 960 $ fps) and obtain the corresponding event sequence. We add $30 \%$ ($30 \%$ is artificially calculated approximate ratio of noise events to effective events in simple real scenes) noisy events with uniform random distribution to the sequence. 

The entire synthetic dataset contains four parts: 
\begin{itemize}
	\item \textbf{HR clear images dataset} consists of $25650$ HR sharp clear frames with various contents, locations, natural and handmade objects, from $270$ video, each of which contains 95 images. It is used as ground truth in training and testing of network. 
	\item \textbf{LR clear images dataset} consists of $25650$ LR sharp clear frames from $270$ video. It is used as ground truth in training and testing of network without SR. 
	\item \textbf{LR blurry images dataset} consists of $25650$ LR blurry noisy frames from $270$ video correspondingly, which simulates the APS frames of the event camera with motion blur and noises. 
	\item \textbf{Event sequences dataset} consists of event sequences corresponding to LR blurry frames. LR blurry images dataset and event sequences dataset are used as inputs of the eSL-Net.
\end{itemize}


According to partitions of GoPro dataset \cite{nah2019ntire}, images and event sequences in the synthetic dataset from $ 240 $ videos are used for training and the rest from $30$ videos for testing.

\section{Experiments}\label{exper}
We train our proposed eSL-Net on the synthetic training dataset for $50$ epoches using NVIDIA Titan-RTX GPUs, and compare it with state-of-the-art event-based intensity reconstruction methods, including EDI \cite{pan2019bringing}, complementary filter method (CF) \cite{scheerlinck2019a} and manifold regularization method (MR) \cite{munda2018}. All methods are evaluated on the synthetic testing dataset and some real scenes \cite{pan2019bringing} to verify intensity reconstruction capability. For the metrics, we use PSNR and SSIM \cite{wang2003multiscale} for quantitative comparison while the visual effect of the reconstructed images for qualitative comparison. In the end, we will test the ability of our method to reconstruct high frame-rate video frames in Section \ref{high-frame}.

\subsection{Intensity Reconstruction Experiments}

Note that EDI, CF and MR can not super resolve the intensity images. Thus for fair comparison, we first replace the HR dictionary $\boldsymbol{D}_{X}$ with a LR dictionary (delete shuffle layers) in eSL-Net to demonstrate the basic ability of intensity reconstruction. Besides, to demonstrate the ability to solve three problems of denoising, deblurring and super resolution simultaneously, the eSL-Net is compared with EDI, CF and MR armed with an excellent SR network RCAN \cite{zhang2018image}.

\input{table1.tex}
\input{table2.tex}
\begin{figure}[htb]	
	\centering	
	\includegraphics[width=12cm]{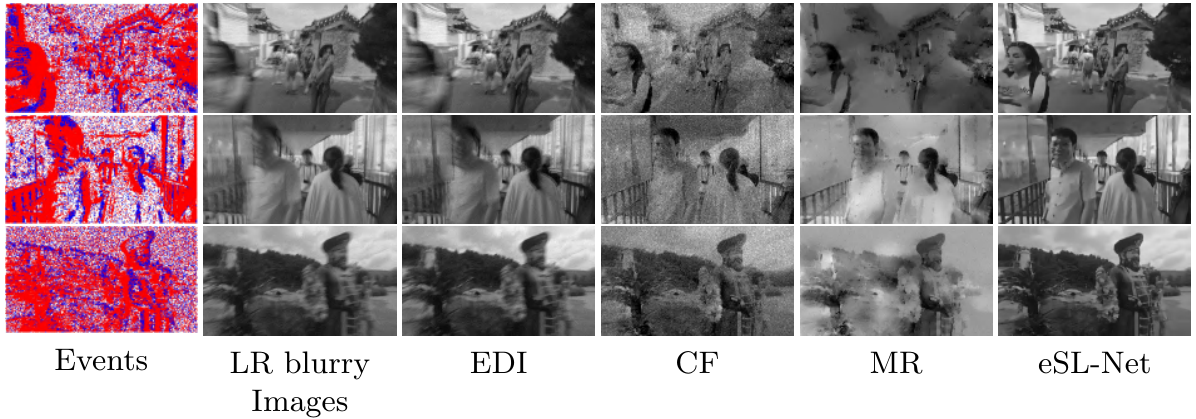}	
	\caption{Qualitative comparison of our outputs without SR to EDI, CF and MR on the synthetic testing dataset.}
	\label{syn_dn}
\end{figure}
\begin{figure}[htb]	
	\centering	
	\includegraphics[width=12cm]{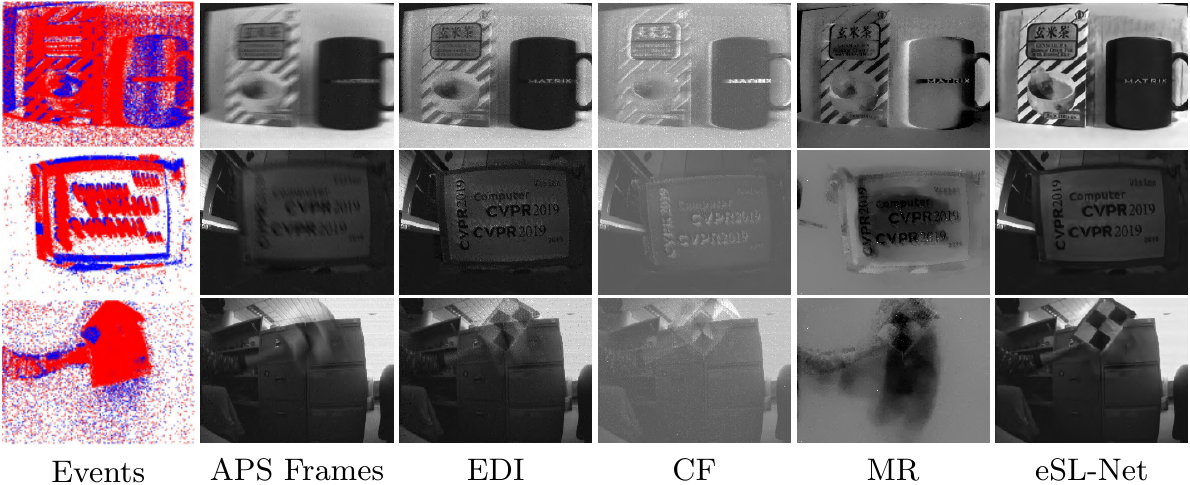}	
	\caption{Qualitative comparison of our outputs without SR to EDI, CF and MR on the real dataset \cite{pan2019bringing}.}
	\label{real_dn}
\end{figure}

For the results without SR, the quantitative comparison is tabulated in Table \ref{dn_table}. One can see that our method outperforms others by a large margin, especially for CF and MR. Note that since results of CF and MR are terrible in the beginning frames of videos, we only choose the middle reconstruction results ($ 16$-$90 $ frames) of each video when calculating PSNR and SSIM. The qualitative results are shown in Fig. \ref{syn_dn} and Fig. \ref{real_dn}. Our method
consistently achieves the best visual performance in terms
of effectively denoising and deblurring.

\begin{figure}[htb]	
	\centering	
	\includegraphics[width=12cm]{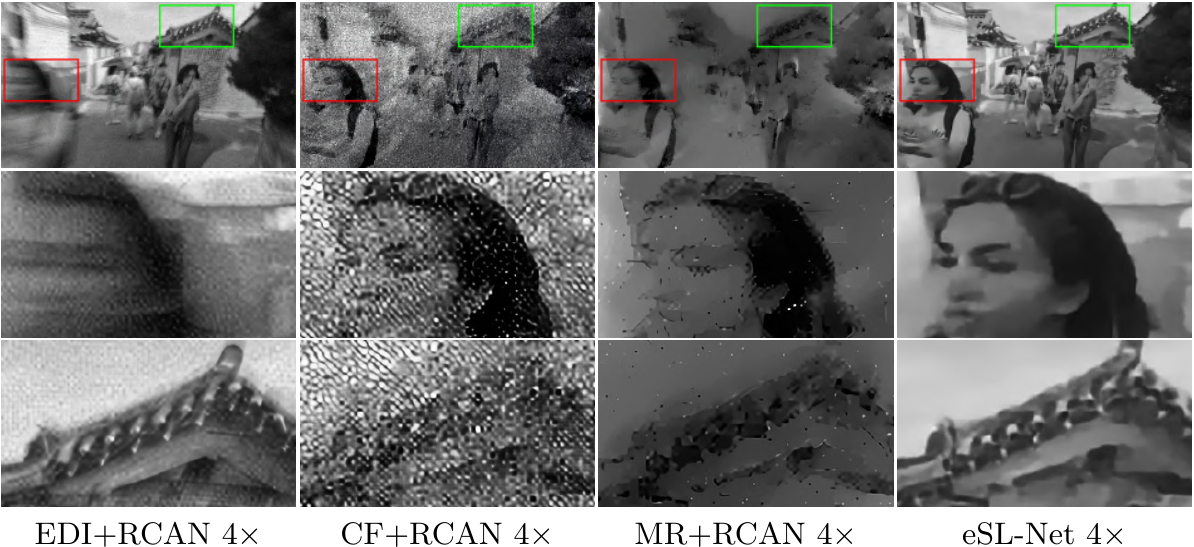}	
	\caption{Qualitative comparison of our outputs to EDI, CF and MR with SR method on the synthetic testing dataset.}
	\label{syn_sr}
\end{figure}
\begin{figure}[!htb]	
	\centering	
	\includegraphics[width=12cm]{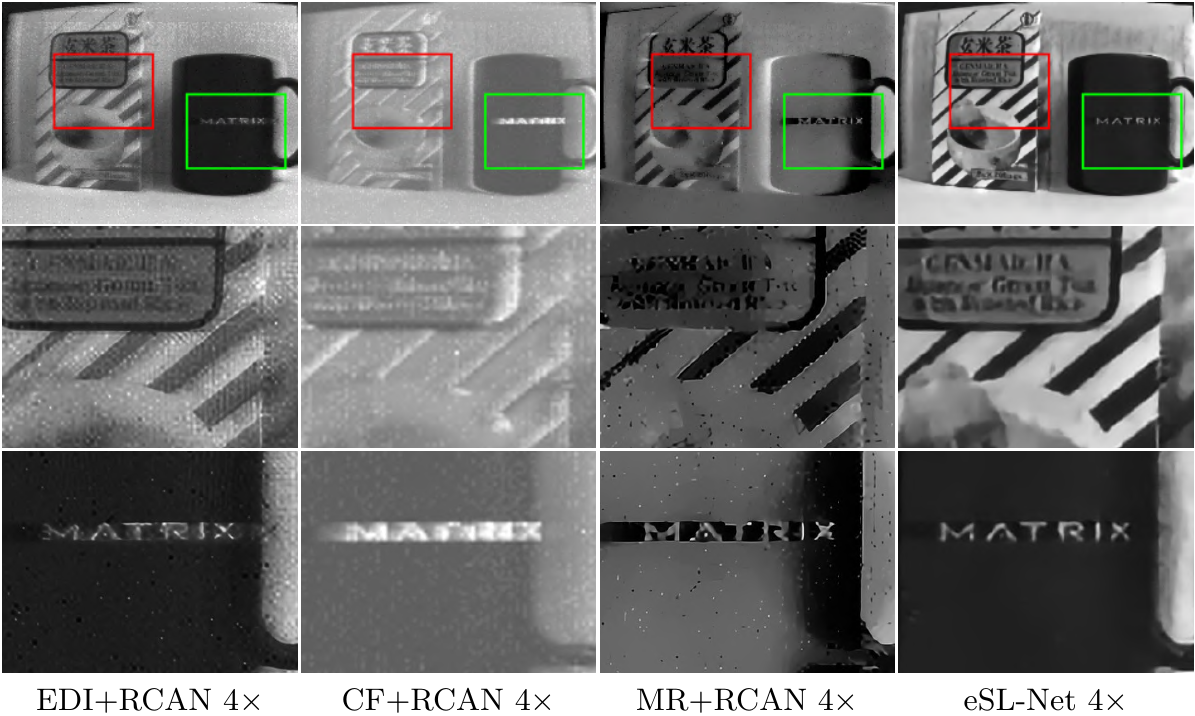}	
	\caption{Qualitative comparison of our outputs to EDI, CF and MR with SR method on the real dataset \cite{pan2019bringing}.}
	\label{real_sr}
\end{figure}

For the results with SR, we show the quantitative results in Table \ref{sr_table}. Compared with Table \ref{dn_table}, all methods become worse when solving the three problems simultaneously. The reason why is that increased resolution aggravates the adverse effects of noise and blur. Even though, our method is still able to maintain a good performance, which means noise, motion blur, low-resolution are well solved simultaneously in our eSL-Net. And the visual comparisons on synthetic and real datasets verifies this inference, as shown in Fig. \ref{syn_sr} and Fig. \ref{real_sr}. Compared to other methods superimposing a complex SR network, our method performs much better on edge recovery and noise removal after super resolution, which proves the superiority of our method.

\subsection{High Frame-Rate Video Experiments}

\begin{figure}[htb]	
	\centering	
	\includegraphics[width=12cm]{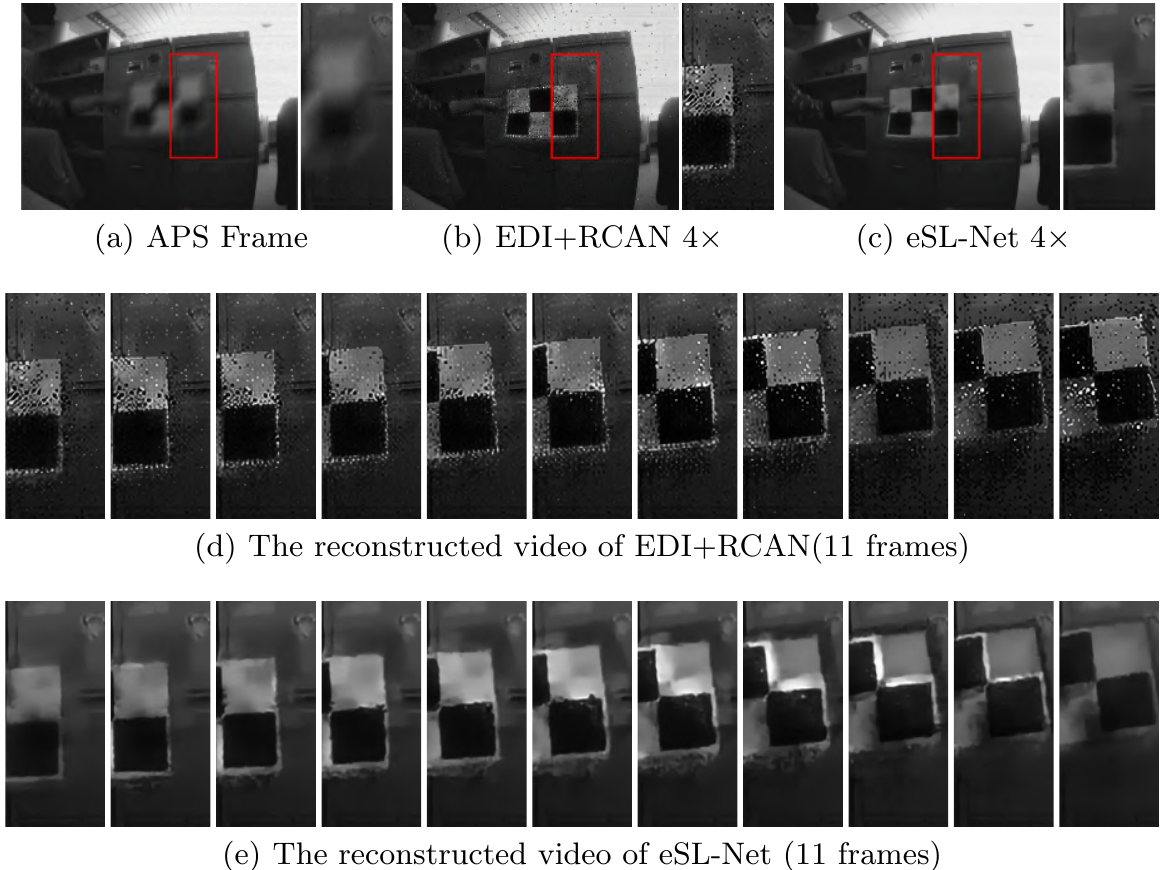}	
	\caption{ An example of the reconstructed result on the real dataset \cite{pan2019bringing}. (a) The LR blurry and noisy input image. On the right is the zoom in of the marked rectangular. (b) The reconstructed frame by EDI with SR of scale 4. (c) The reconstructed frame by eSL-Net with SR of scale 4. (d)-(e) show selected 11 frames of the reconstructed video through EDI+RCAN and eSL-Net from (a) and an event sequence respectively.}
	\label{mul_image}
\end{figure}

Theoretically, we can generate a video with frame-rate as high as the DVS's eps. However, when the frame-rate is too high, there are only a few events in the interval of two continuous frames, which results in a subtle difference between the reconstructed two frames, and even difficulty to distinguish them. Therefore, in order to balance the frame-rate and the visual effect, in our experiments, our method generates coutinuous frames with frame-rate $ 21 $ times higher than the original APS frames. To show the changes among frames more clearly, 11 frames are taken at equal intervals from continuous frames recovered from an image and an event sequence by eSL-Net, as shown in Fig. \ref{mul_image} (e), which demonstrates the effectiveness of our method on HR and high-frame-rate video reconstruction. Without retraining, our method is able to generate continuous frames while preserving more realistic and richer details.

Fig. \ref{mul_image} shows qualitative comparisons between our method and EDI+RCAN about high frame-rate video reconstruction on real dataset. Obviously, our method is superior in edge recovery and noise removal.

\section{Conclusion}
 In this paper, we proposed a novel network named \textbf{eSL-Net} for high-quality image reconstruction from event cameras. Enhanced by events, the degeneration model of event cameras can easily cope with the motion blur, noises and low spatial resolution problems. Particularly, exploiting sparse learning framework, we proposed an explainable network, i.e. eSL-Net, which is obtained by unfolding the iterative soft thresholding algorithm. Besides, the eSL-Net can be easily extended to generate high frame-rate videos only by a simple transform of the event steam. Experiments on synthetic and real-world data demonstrate the effectiveness and superiority of our eSL-Net.

\section*{Acknowledge} 
The research was partially supported by the National Natural Science Foundation of China under Grants 61871297. And the research was partially supported by the Fundamental Research Funds for the Central Universities.



\clearpage
%
%
\bibliographystyle{splncs04}
\bibliography{eccv1819}
\end{document}

%% file: table1.tex
\begin{table}[!t]
	\centering
	\caption{Quantitative comparison of our outputs without SR to EDI, CF and MR on the synthetic testing dataset.}  
	\begin{tabularx}{12cm}{p{1.6cm}<{\centering} p{2.5cm}<{\centering} p{2.5cm}<{\centering} p{2.5cm}<{\centering} p{2.5cm}<{\centering} }
		\hline  
		Methods & EDI  & CF & MR & eSL-Net \\  
		\hline  
		PSNR(dB)  & 22.42 & 19.75 & 13.76 & \textbf{30.23} \\  
		SSIM  & 0.6228 & 0.4131 & 0.4960 & \textbf{0.8703}\\  
		\hline  
	\end{tabularx} 
	\label{dn_table} 
\end{table}

%% file: table2.tex
\begin{table} [!t]
	\centering
	\caption{Quantitative comparison of our outputs to EDI, CF and MR with SR method on the synthetic testing dataset.}  
	\begin{tabularx}{12cm}{p{1.6cm}<{\centering} p{2.5cm}<{\centering} p{2.5cm}<{\centering} p{2.5cm}<{\centering} p{2.5cm}<{\centering} }  
		\hline  
		Methods & EDI+RCAN 4$\times$  & CF+RCAN 4$\times$ & MR+RCAN 4$\times$ & eSL-Net 4$\times$ \\  
		\hline  
		PSNR(dB)  & 12.88 & 12.89 & 12.89 & \textbf{25.41} \\  
		SSIM  & 0.4647 & 0.4638 & 0.4643 & \textbf{0.6727}\\  
		\hline  
	\end{tabularx} 
	\label{sr_table} 
\end{table}